\documentclass[10pt,twocolumn,letterpaper]{article}

\usepackage{wacv}
\usepackage{times}
\usepackage{epsfig}
\usepackage{graphicx}
\usepackage{amsmath}
\usepackage{amssymb}
\usepackage{enumitem}

\usepackage{xcolor}

\usepackage{multirow}
\usepackage{soul}

\def\wacvPaperID{1352} 

\wacvfinalcopy 

\usepackage[breaklinks=true,bookmarks=false]{hyperref}

\ifwacvfinal\fi
\begin{document}

\title{Revisiting Batch Normalization for Improving Corruption Robustness}

\author{Philipp Benz\footnote{Equal contribution}\\
{\tt\small pbenz@kaist.ac.kr}
\and
Chaoning Zhang$^*$\\
{\tt\small chaoningzhang1990@gmail.com}
\and
Adil Karjauv \\
\and
In So Kweon\\
\and
\\Korea Advanced Institute of Science and Technology (KAIST)\\
{\tt\small $^*$ Equal Contribution}
}

\maketitle
\thispagestyle{empty}

\begin{abstract}
The performance of DNNs trained on clean images has been shown to decrease when the test images have common corruptions. In this work, we interpret corruption robustness as a domain shift and propose to rectify batch normalization (BN) statistics for improving model robustness. This is motivated by perceiving the shift from the clean domain to the corruption domain as a style shift that is represented by the BN statistics. We find that simply estimating and adapting the BN statistics on a few (32 for instance) representation samples, without retraining the model, improves the corruption robustness by a large margin on several benchmark datasets with a wide range of model architectures. For example, on ImageNet-C, statistics adaptation improves the top1 accuracy of ResNet50 from 39.2\% to 48.7\%. Moreover, we find that this technique can further improve state-of-the-art robust models from 58.1\% to 63.3\%.
\end{abstract}

\section{Introduction}
In the past few years, deep learning has shown unprecedented performance in various vision tasks~\cite{he2016deep,huang2017densely,girshick2015fast,ren2015faster,long2015fully,zhang2019revisiting,zhang2020deepptz,feng2020adversarial,kim2019recurrent,pan2020unsupervised,kim2020video,zhang2020resnet,kim2020Devil}. However, models are still widely known to be vulnerable to adversarial examples ~\cite{szegedy2013intriguing,goodfellow2014explaining}, leading to significant attention on adversarial robustness~\cite{benz2020data,zhang2019cd-uap,zhang2020understanding,benz2020double,bai2020targeted,benz2020robustness}. In practice, common corruptions~\cite{hendrycks2019benchmarking}, such as Gaussian noise, have also shown to decrease the performance non-trivially, causing critical concerns for the need to improve corruption robustness.

\begin{figure}[!htbp]
    \centering
    \includegraphics[width=0.7\linewidth]{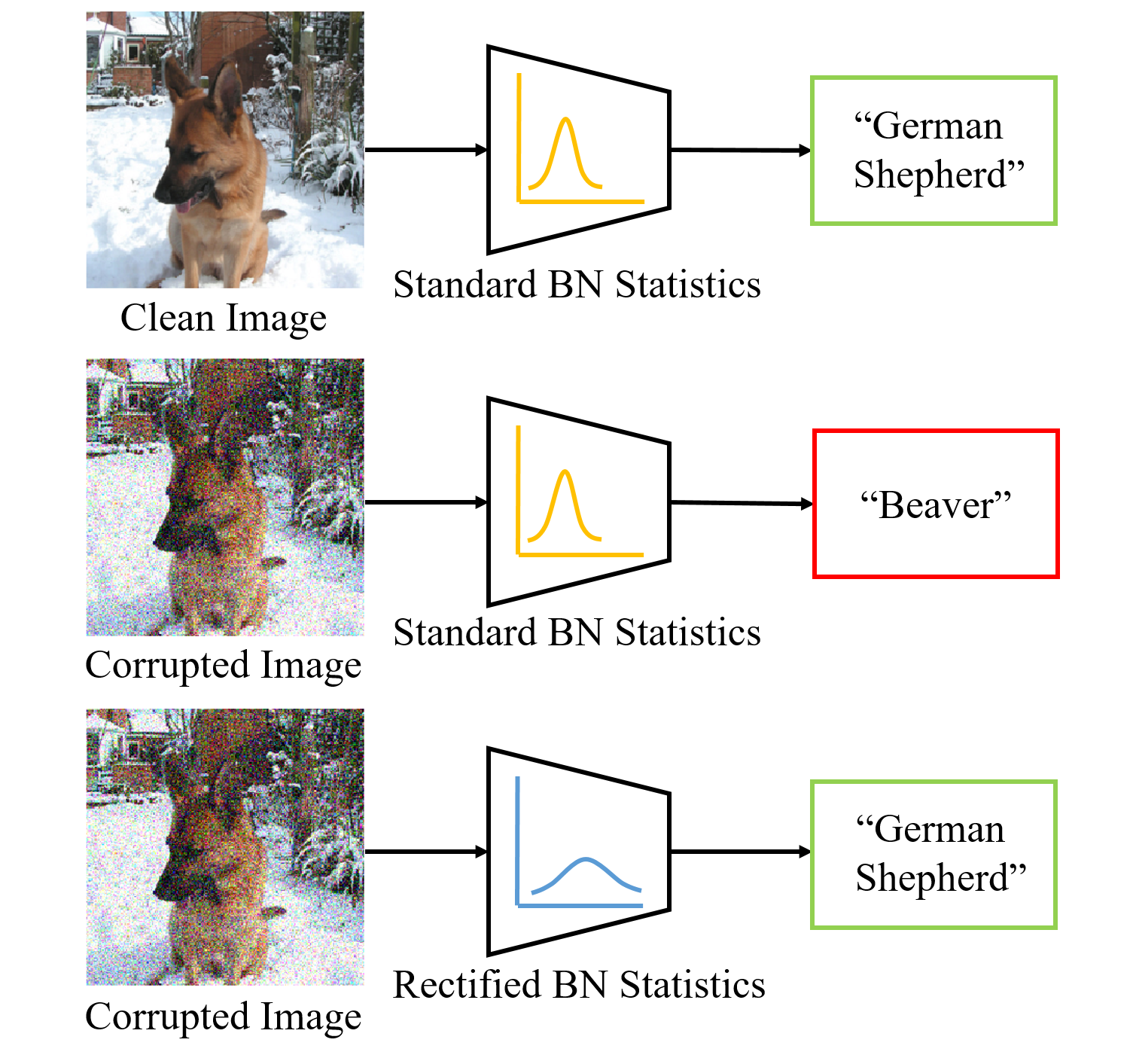}
    \caption{Improving corruption robustness by rectifying the BN statistics. An image under corruption changes the prediction from ``German Shepherd" to ``Beaver". After rectifying the BN statistics, the corrupted image is classified correctly.}
    \label{fig:teaser}
\end{figure}

In a parallel line of research on domain adaptation~\cite{ben2007analysis,daume2009frustratingly,ganin2015unsupervised,ganin2016domain,tzeng2017adversarial}, an unlabeled target domain dataset is exploited to improve the model generalization capability to a target domain. Roughly speaking, the clean images and corrupted images can be seen as coming from different domains: clean (source) domain and corruption (target) domain, respectively. Despite this conceptual similarity between domain adaptation and corruption robustness, the community tends to treat them as two distinct issues: Domain adaptation usually has a predefined target domain with unlabeled dataset~\cite{ganin2015unsupervised,ganin2016domain}, while corruption robustness normally does not assume such a predefined corruption type~\cite{hendrycks2019augmix}. One straightforward solution to improve the corruption robustness is performing data-augmentation with corrupted images during training. The limitation of this solution is that a model trained on images of a certain corruption type might be vulnerable to another type of corruption. For example, it has been shown that a model trained with Gaussian noise augmentation increases the robustness against Gaussian noise, which is expected while decreasing the robustness against contrast and fog corruptions~\cite{ford2019adversarial,yin2019fourier}.

Given the constraint that the corruption type is unknown during the training stage, we can still exploit the corruption type during the inference stage. It is reasonable to assume that the corruption variant will not change for a short period during inference. For example, in autonomous driving, the weather condition is highly likely to be stable at least in a short period in most cases, thus, the system can capture a stream of unlabeled images to represent the current weather condition.
With a few representation samples, it is meaningless as well as impractical to directly apply the general domain adaptation techniques for retraining to improve robustness. In domain adaptation, there is a line of work adapting the feature statistics instead of adapting features~\cite{sun2017correlation,cariucci2017autodial,li2016revisiting,romijnders2019domain}. Among them, what is most applicable in the context of corruption robustness is adaptive batch normalization (AdaBN) which simply adapts the batch normalization statistics without the need to retrain the model~\cite{li2016revisiting}.

BN~\cite{ioffe2015batch} has been widely adopted in modern DNNs, for example, most (if not all) seminal classification models, such as ResNet~\cite{he2016deep}, DenseNet~\cite{huang2017densely}, ResNeXt~\cite{xie2017aggregated}, use BN by default. Moving average is often applied over the training dataset to estimate the population statistics for inference~\cite{ioffe2015batch}. This inevitably causes a shift in the statistics if the test sample is from a corruption domain different from the domain where the statistics were estimated. As indicated in Figure~\ref{fig:teaser}, we investigate and find that such influence on the model performance can be at least partially mitigated by estimating and adapting the statistics with a few representation samples from the corruption domain. Our investigation suggests that the model robustness against corruptions can be significantly improved by applying this simple yet effective technique. Concurrent to our work,~\cite{schneider2020improving} also shows that updating the BN statistics during inference can improve the robustness against common corruptions. Their work is motivated by reducing the covariate shift, while this work approaches the problem from a style vs.\ content perspective.

Our contributions are summarized as follows:
\begin{itemize}[noitemsep, topsep=0pt]
    \item We interpret corruption robustness as a domain adaptation problem, inspired by which we investigate the effectiveness of adapting BN statistics on model corruption robustness.
    \item On several benchmark datasets, including CIFAR10-C, CIAR100-C, ImageNet-C, we demonstrate that simply adapting BN statistics can significantly improve the model corruption robustness.
    \item We show that the technique is also orthogonal to current SOTA methods that improve the corruption robustness. For example, the accuracy of the SOTA method ``DeepAugment + AugMix" can be improved from $58.1\%$ to $63.3\%$.
\end{itemize}

\section{Related works}
\subsection{Batch Normalization} BN was originally introduced to reduce the covariate shift for faster convergence~\cite{ioffe2015batch}. It has been found in~\cite{santurkar2018does} that BN leads to fast convergence due to the smoothed optimization landscape. One recent work~\cite{awais2020revisiting} revisits the internal covariate shift and argues that it is crucial to understand how BN works. The success of BN has also been found to be connected to the decoupling of length and direction~\cite{kohler2019exponential}. Even though the mechanism of how BN helps training remains not fully clear, the phenomenon that BN boosts the performance and convergence is empirically proven in a wide range of works~\cite{he2016deep,huang2017densely}. BN also has a regularization effect due to the mini-batch stochasticity and increases the model generalization capability~\cite{luo2018towards}. Recently, BN has also been explored for its impact on adversarial robustness~\cite{benz2020batch,xie2019intriguing,awais2020towards}. For example, BN has been found to increase standad accuracy while at the cost of adversarial robustness~\cite{benz2020batch} and the reason is attributed to BN shifting the models to rely more on non-robust features. 

A large batch size is required for BN, which limits its applications. Layer normalization~\cite{ba2016layer} addresses this issue by exploiting the channel dimension instead of the batch dimension. For the purpose of style transfer, instance normalization (IN)~\cite{vedaldi2016instance}, which only performs normalization on the individual feature channel, has also been explored. Inspired by the interpretation in~\cite{huang2017arbitrary} that IN performs a form of style normalization, Batch-Instance normalization has been proposed in~\cite{nam2018batch} for automatically learning to normalize only disturbing styles while preserving useful styles. More recently, Group normalization (GN) has been proposed to perform normalization with groups of channels~\cite{wu2018group}.

\subsection{From domain shift to corruption robustness}
In practice, the distribution shift occurs as a major concern~\cite{ganin2015unsupervised,ford2019adversarial}. To address this concern, numerous works assume having access to unlabeled samples from the target domain and bridge the gap between the source domain and target domain by applying the techniques of domain adaptation~\cite{ganin2015unsupervised,ganin2016domain,tzeng2017adversarial,hoffman2017cycada,hong2018conditional}. Another line of work focuses on the model robustness to common corruptions~\cite{hendrycks2019benchmarking}, which can also be seen as domain shift, \ie gap between clean domain and corruption domain. Hendrycks \etal established rigorous benchmarks for image classifier robustness by introducing ImageNet-C which is a variant of ImageNet with common corruptions~\cite{hendrycks2019benchmarking}.
Benchmarking robustness on other applications~\cite{michaelis2019benchmarking,kamann2020benchmarking} has also been proposed, demonstrating the community's interest in corruption robustness. To improve corruption robustness, data augmentation can be one straightforward solution, however, the augmented model can not be generalized to other corruptions. For example, ~\cite{ford2019adversarial} has shown that augmentation with Gaussian noise improves the model robustness against Gaussian noise while reducing the robustness against contrast and fog corruptions~\cite{ford2019adversarial}. Training on images with transformed style has been found to improve the corruption robustness~\cite{geirhos2018imagenet}. AugMix has been proposed in~\cite{hendrycks2019augmix} as a simple prepossessing method combining consistency loss for improving robustness to unseen corruptions. Recently, DeepAugment has been proposed~\cite{hendrycks2020many} and combined with AugMix achieves SOTA corruption robustness.

\subsection{Aligning or Adapting feature statistics}
The above methods of improving robustness to common corruptions often require training the model on a special dataset or adopting a specially designed augmentation technique. In domain adaptation, aligning or adapting feature normalization statistics, \ie mean and variance, has been found beneficial for bridging the gap between the source domain and target domain~\cite{sun2017correlation,cariucci2017autodial}.
Adaptive Batch Normalization (AdaBN), has been proposed in~\cite{li2016revisiting} showing that adapting the statistics with target domain images improves the performance on the target domain. In this work, we explore the effectiveness of adapting BN statistic with a few representation corruption samples to improve the corruption robustness. Somewhat surprisingly, we find that this simple technique can improve the corruption robustness by a significant margin.

\section{Background and motivation}
\subsection{Revisiting classical batch normalization}
We briefly summarize how BN works in practice. For a certain layer in the DNN, the feature layers of a mini-batch are represented by $\mathcal{B} = \{x_1, ..., x_m\}$. During training, BN performs normalization on this mini-batch as follows,
\begin{equation}
     \hat{x}_i = \frac{x_i - \mu_\mathcal{B}}{\sigma_\mathcal{B}} \cdot \gamma + \beta
     \label{BN_principle}
\end{equation}
where $\gamma$ and $\beta$ denote the learnable parameters scale and shift, respectively. In the remainder of this paper, we ignore $\gamma$ and $\beta$ for simplicity.
For 2D images, the mean $\mu_\mathcal{B}$ and variance $\sigma^2_\mathcal{B}$ for a feature layer $x_i$ of spatial width $W$ and height $H$, are calculated as:
\begin{equation}
\centering
\begin{aligned}
\label{eq:mub_sigmab}
     \mu_\mathcal{B} = \frac{1}{M} \frac{1}{W} \frac{1}{H} \sum_{i=1}^{M} \sum_{j=1}^{W} \sum_{q=1}^{H} x_i^{jq},\\ \sigma_\mathcal{B}^2 = \frac{1}{M} \frac{1}{W} \frac{1}{H} \sum_{i=1}^{M} \sum_{j=1}^{W} \sum_{q=1}^{H} (x_i^{jq}-\mu_\mathcal{B})^2.
\end{aligned} 
\end{equation} 
where $j$ and $q$ indicate the spatial position of the feature layer.
BN works in different modes during training and test stage. During the training stage, the normalization depends on the mini-batch statistics to ensure stable training, while this dependency becomes unnecessary during the test stage. Thus, the population statistics are adopted to make the inference depend on the individual input in a deterministic manner. Empirically, this population statistics $\mu_\mathcal{P}$ and $\sigma^2_\mathcal{P}$ are estimated over the whole training dataset through moving average. It is worth mentioning that $\mu_\mathcal{B}$ and $\sigma^2_\mathcal{B}$ are almost the same as $\mu_\mathcal{P}$ and $\sigma^2_\mathcal{P}$, thus, in general, there is no mismatch in the BN statistics during training and testing.

\subsection{BN statistics: Style instead of content}
\begin{figure}[!htbp]
    \centering
    \includegraphics[width=\linewidth]{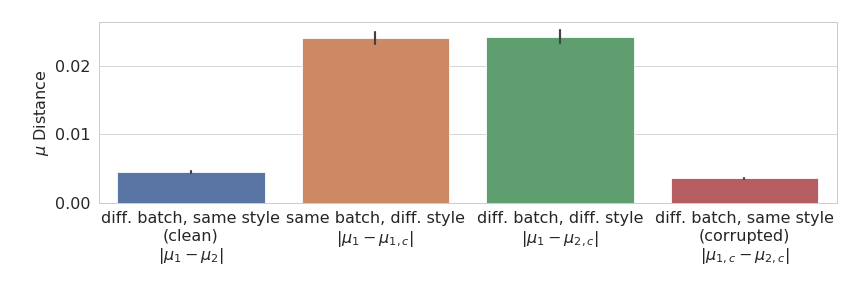}
    \bigskip
    \includegraphics[width=\linewidth]{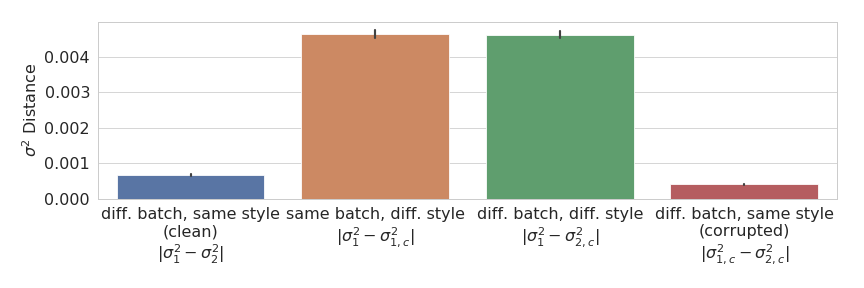}
    \caption{Absolute distance between the mean (top) and variance (bottom) for either the same or different batches of clean or corrupted samples. The results are averaged over $100$ measurements. The statistics were calculated for a randomly selected layer of ResNet50 pretrained on ImageNet.}
    \label{fig:style_vs_content}
\end{figure}

The information of an image can be described through content and style information~\cite{gatys2016image}.
Instance Normalization (IN)~\cite{ulyanov2016instance} was introduced to discard instance-specific contrast information from an image during style transfer. For a feature layer its individual mean and variance can be computed as:
\begin{equation}
     \mu_i = \frac{1}{W} \frac{1}{H} \sum_{j=1}^{W} \sum_{q=1}^{H} x_i^{jq}, \quad \sigma_i^2 = \frac{1}{W} \frac{1}{H} \sum_{j=1}^{W} \sum_{q=1}^{H} (x_i^{jq}-\mu_i)^2.
\end{equation} 
According to~\cite{huang2017arbitrary}, $FC_i = \frac{x_i - \mu_i}{\sigma_i}$ indicate the feature content inherent to the sample by performing a form of style normalization, namely $\mu_i$ and $\sigma^2_i$. It has been shown in~\cite{huang2017arbitrary} that simply adjusting the mean and variance of a generator network can control the style of the generated images.
BN normalizes feature statistics for a batch of samples instead of a single sample. Thus, BN can be intuitively understood as normalizing a batch of samples with different contents to be centred around a single style. With this understanding, the population statistics $\mu_\mathcal{P}$ and $\sigma^2_\mathcal{P}$ represent the style information instead of the content information in $x_i$. 
To verify this hypothesis, we measure the absolute difference for BN statistics under different network inputs. The BN statistics are calculated for a randomly selected layer.
The statistics are either calculated for samples from the ImageNet test dataset (indicated by $\mu$, $\sigma^2$), or its corruption variants corrupted through Gaussian noise (indicated by  $\mu_c$, $\sigma^{2}_c$). We averaged the results over $100$ batches and present them in Figure~\ref{fig:style_vs_content}.
Comparing different batches coming from the same distribution (either corrupted or uncorrupted) we observe that the BN statistics are very similar and do not deviate much, indicating that these batches indeed have similar styles despite different content. 
Comparing batches of clean samples and corrupted samples, a relatively greater difference can be observed in the BN statistics for the same or different content. 
Overall, these results suggest that BN statistics are mainly determined by the mini-batch style instead of their content.

\subsection{Motivation for rectifying batch normalization}
To motivate our approach, we first showcase the influence of input corruptions on the BN statistics. To measure the shift caused by corruptions, we treat each feature output as a vector~\cite{zhang2020understanding} and adopt the cosine similarity $cos$ between the feature output of a clean batch and a corrupted one and finally average over the batch size. 
The more similar two feature layer outputs, the more close the $cos$ value is to $1$~\cite{zhang2020understanding}. A value of $0$ indicates that the two feature outputs are maximally dissimilar.

\begin{figure}[!htbp]
    \centering
    \includegraphics[width=0.9\linewidth]{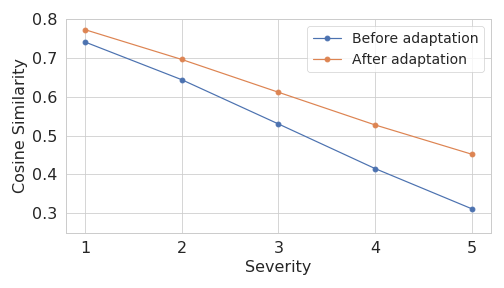}
    \caption{Cosine similarity between different feature outputs for a ResNet50 (ImageNet). The blue line indicates the cosine similarity between feature outputs of clean and corrupted images evaluated on the not yet rectified model. The orange line shows the similarity between feature outputs of clean images and feature output of corrupted images evaluated on the rectified model.}
    \label{fig:sim_cl_vs_corr_feature_maps_before_after_adaptation}
\end{figure}

In Figure~\ref{fig:sim_cl_vs_corr_feature_maps_before_after_adaptation}, we visualize the cosine similarity for a standard model over $5$ severity levels of Gaussian noise corruption (blue line). With increasing severity the cosine similarity decreases indicating a greater deviation of the two feature layer outputs.
To demonstrate that this degradation can be mitigated by rectification of the BN statistics, we compute the cosine similarity between the feature layer output of the original model and a rectified model for corrupted input samples (see Figure~\ref{fig:sim_cl_vs_corr_feature_maps_before_after_adaptation} orange line). 
We observe that rectifying the BN statistics improves the cosine similarity values over all severity values. 
The results support our hypothesis that the performance degradation caused by corruptions can be attributed to the shifted style information induced through the corruptions.
This observation motivates us to rectify the BN statistic with a small number of samples to improve model robustness under corruptions. 

In practice, it is not challenging to obtain a reasonably small number of representation samples. For example, in the scenario of autonomous driving, weather conditions can change from day to day but tend to be consistent over a shorter time-frame. Thus a system can capture a few images (without labels) after a significant change in the conditions. 

\section{Experimental Setup}
ImageNet-C was proposed by~\cite{hendrycks2019benchmarking} to benchmark neural network robustness against common corruptions. ImageNet-C has the same image content as that of the ImageNet validation dataset (1000 classes and 50 images for each class) but perturbed with various corruptions. 
Specifically, there are 15 test corruptions and another four hold-out corruptions. 
Similar to~\cite{hendrycks2019augmix}, we evaluate on the 15 test corruptions. Each corruption type has 5 different severities. The same authors proposed similar datasets for CIFAR10 and CIFAR100, termed CIFAR10-C and CIFAR100-C respectively. 

To rectify the BN statistics we randomly select a batch of $32$ representation samples from the corruption dataset of the respective severity. We calculate $\mu_{\mathcal{B}}$ and $\sigma^2_{\mathcal{B}}$ according to Eq.~\ref{eq:mub_sigmab} and update the population statistics with them. 

We evaluate the performance of rectifying the BN statistics on various models trained on the corresponding clean dataset. A wide range of state-of-the-art models is adopted for evaluation. Following Hendrycks~\etal we adapt the corruption error ($CE$) as a metric 
\begin{equation}
CE^f_c = \frac{\sum_{s=1}^5 E_{s,c}^f}{\sum_{s=1}^5 E_{s,c}^\text{AlexNet}},
\end{equation}
where $E_{s,c}^f$ indicates the error for model $f$ evaluated under corruption $c$ for severity $s$.
The mean over the different corruption types results in the mean CE indicated by mCE from here on. Additionally, we report the top1 accuracy (Acc), also averaged over the different corruption types. We indicate the metrics after adaptation with an asterisk, \ie mCE*, and Acc* to differentiate from that before rectification. For the Acc metric the higher the better, while for the mCE the lower the better. 

\section{Experimental Results}
\subsection{Evaluation on CIFAR10-C and CIFAR100-C}
First, we provide evidence for the effectiveness of our proposed method rectifying the BN statistics to increase corruption robustness on the CIFAR10-C and CIFAR100-C benchmark datasets. To showcase the general applicability of our approach we rectify the BN statistics on a variety of models. 
The results for CIFAR10-C and CIFAR100-C are reported in Table~\ref{tab:cifar10c} and Table~\ref{tab:cifar100c}, respectively.
For both datasets rectifying the BN statistics improves the robustness by a significant margin across all tested models. For CIFAR10-C, the accuracy of most models is improved by around $10\%$ points. The lowest performance increase is observed on ResNet-20, which is still around $5\%$ points. Notably, ResNet-20 also exhibits the lowest initial robustness with an mCE of $106.4\%$ indicating being less robust to corruptions than AlexNet, but with an mCE of $90.7\%$ after rectification.

In the case of CIFAR100-C, a trend can be observed that the models with higher capacity exhibit a higher initial accuracy and also show a higher performance increase after rectification of the BN statistics. In particular, WRN-28-10, ResNeXt-29, and DenseNet all enjoy a robust performance increase of more than $12\%$ points while the relatively smaller ResNet-20, ResNet-56, and VGG-19 increase by around $5\%$ points.

Overall, the CIFAR-C results suggest that rectifying the BN statistics results in a minimum accuracy increase of $5\%$ points but more often $10\%$ and higher. This suggests that rectification of BN statistics is a simple yet effective technique to boost model robustness against common corruptions.

\begin{table}[!htpb]
\centering
\caption{Evaluation results on CIFAR10-C.}
    \scalebox{0.9}{
\label{tab:cifar10c}
    \begin{tabular}{c|cc|cc}
        \hline
        Model & Acc & Acc* & mCE & mCE* \\ \hline
        ResNet-20    &  68.2  & 73.0 & 106.4 & 90.7  \\ 
        ResNet-56    &  70.7  & 81.4 & 98.5  & 63.0  \\ 
        ResNet-18    &  73.9  & 84.3 & 87.9  & 53.0  \\ 
        ResNet-50    &  74.0  & 83.1 & 87.4  & 57.6  \\ 
        VGG-19       &  72.9  & 81.0 & 90.1  & 63.7  \\ 
        WRN-28-10    &  78.4  & 86.8 & 73.1  & 44.5  \\ 
        ResNeXt-29   &  75.0  & 85.5 & 85.1  & 49.9  \\ 
        DenseNet     &  76.7  & 87.6 & 80.3  & 42.4  \\ 
        \hline
    \end{tabular}
    }
\end{table}

\begin{table}[!htpb]
\centering
\caption{Evaluation results on CIFAR100-C}
\scalebox{0.9}{
\label{tab:cifar100c}
    \begin{tabular}{c|cc|cc}
        \hline
        Model & Acc & Acc* & mCE & mCE* \\ \hline
        ResNet-20   &  38.9  & 44.5 & 96.1  & 87.4  \\ 
        ResNet-56   &  43.8  & 48.0 & 88.5  & 81.9  \\ 
        VGG-19      &  45.3  & 51.4 & 86.1  & 76.6  \\ 
        WRN-28-10   &  53.0  & 65.3 & 74.1  & 54.8  \\ 
        ResNeXt-29  &  52.7  & 66.0 & 74.6  & 53.7  \\ 
        DenseNet    &  52.9  & 65.8 & 74.5  & 54.2  \\ 
        \hline
    \end{tabular}
    }
\end{table}

\subsection{Evaluation on ImageNet-C}
Besides CIFAR, ImageNet is another commonly used benchmark-dataset to evaluate classification accuracy. As above, we adopt its corrupted version ImageNet-C to evaluate the performance of different benchmark models obtained from the \texttt{torchvision} repository. The results are presented in Table~\ref{tab:imagenetc}. 
The results show that a change in the BN statistics can also result in significant performance improvements of up to $10\%$ points.
Similar to the trend observed on CIFAR100-C, we note a trend that models of relatively higher-capacity (Wide Resnet, ResNeXt, and DenseNet) exhibit a higher initial accuracy (higher than $46\%$) compared to the relatively smaller models. However, opposed to the trend on CIFAR100-C, the performance improvement on the relatively smaller models is slightly larger.

\begin{table}[!htpb]
\centering
\caption{Evaluation results on ImageNet-C with the pretrained models provided in \texttt{torchvision}.}
\scalebox{0.9}{
    \label{tab:imagenetc}
    \begin{tabular}{c|cc|cc}
    \hline
    Model & Acc & Acc* & mCE & mCE* \\ \hline
    VGG-19 (BN)                & 35.4 & 45.5 & 81.6 & 69.3  \\
    ResNet 18                  & 32.9 & 41.8 & 84.7 & 73.7  \\
    ResNet 50                  & 39.2 & 48.7 & 76.7 & 64.9  \\
    Wide ResNet 101            & 46.3 & 51.7 & 67.7 & 60.9  \\
    ResNeXt 101                & 47.1 & 53.9 & 66.7 & 58.2  \\
    DenseNet 161               & 47.6 & 54.7 & 66.4 & 57.4  \\
    \hline
    \end{tabular}
}
\end{table}

\subsection{Evaluation on state-of-the-art models}
The above pretrained models are not optimized for improving the model robustness against common corruptions. We further test whether a similar performance boost can be observed on models that are optimized for achieving state-of-the-art robustness. AugMix was proposed in~\cite{hendrycks2019augmix} as a simple preprocessing method together with a consistency loss. Despite simplicity, it achieves competitive robustness against corruptions, outperforming other approaches by a large margin. More recently, it was proposed to strengthen AugMix by combining it with DeepAugment, achieving state-of-the-art performance~\cite{hendrycks2020many}. The adopted model architecture is ResNet50. The comparison results are shown in Table~\ref{tab:sota_models}. Compared with the baseline (vanilla ResNet50), training with ``AugMix" and ``DeepAugment + AugMix" improves the corruption robustness by a large margin. Strikingly, we observe that adapting the BN statistics also improves the accuracy from $48.3\%$ to $56.7\%$ for ``AugMix". For the SOTA training method ``DeepAugment + AugMix", adapting the BN statistics can still non-trivially improve the accuracy from $58.1\%$ to $63.3\%$. Similar robustness can also be observed for the metric of mCE.

\begin{table}[!htpb]
\centering
\caption{Evaluation results on ImageNet-C with state-of-the-art models.}
\scalebox{0.9}{
\label{tab:sota_models}
\begin{tabular}{c|cc|cc}
\hline
Model & Acc & Acc* & mCE & mCE* \\ \hline
Baseline                   & 39.2 & 48.7 & 76.7 & 64.9  \\
Augmix                     & 48.3 & 56.7 & 65.3 & 55.0  \\ 
DeepAugment + Augmix          & 58.1 & 63.3 & 53.6 & 47.0  \\
\hline
\end{tabular}
}
\end{table}

\subsection{Evaluation on adversarially trained models}
We further evaluate whether adversarially trained models~\cite{goodfellow2014explaining,madry2017towards} can also benefit from rectifying the BN statistics. 
For evaluation, we use the publicly available robust ResNet-50 models for CIFAR10 and ImageNet from~\cite{robustness}. The models were adversarially trained with adversarial examples either bounded through an L$_2$ or L$_\infty$ norm with an upper bound of $\epsilon$ for a pixel range in $[0,1]$. For the robustified CIFAR10 models, it can be observed that adversarial training alone already improves the initial robustness significantly. The two adversarial robust models achieve the highest accuracy among all CIFAR10 models. Rectifying the BN statistics additionally increases the corruption robustness. The performance increase on the robust CIFAR models still is around $3\%$ points, which is less compared to that observed on the standard CIFAR models. 
For the adversarially trained ImageNet models, we observe the opposite trend. For the scenario without BN statistics rectification, adversarially trained ResNet-50 models exhibit a lower corruption accuracy than the normal models. However, after the BN statistics rectification, the corruption accuracy increases by about $15\%$, which is more than that observed on adversarially trained CIFAR10 models.

\begin{table}[t]
\centering
\caption{Evaluation results on CIFAR10-C (top) and ImageNet-C (bottom) on adversarially trained models.}
\setlength\tabcolsep{4.0pt}
\scalebox{0.9}{
\label{tab:adversarial_trained_models}
    \begin{tabular}{c|cc|cc}
        \hline
        Model & Acc & Acc* & mCE & mCE*       \\ \hline
        ResNet-50 (L$_2$, $\epsilon=0.5$)        & 83.6 & 86.6 & 50.8 & 43.8 \\
        ResNet-50 (L$_\infty$, $\epsilon=8/255$) & 79.2 & 82.3 & 64.5 & 57.6 \\
        \hline 
        ResNet-50 (L$_2$, $\epsilon=3.0$)        & 31.1 & 46.9 & 87.3 & 68.4 \\
        ResNet-50 (L$_\infty$, $\epsilon=8/255$) & 23.8 & 38.1 & 96.7 & 79.7 \\
        \hline
    \end{tabular}
    }
\end{table}

\section{Analysis and Discussion}
\subsection{Number of representation samples}
\begin{figure}[!htbp]
    \centering
    \parbox{.48\linewidth}{
    \includegraphics[width=\linewidth]{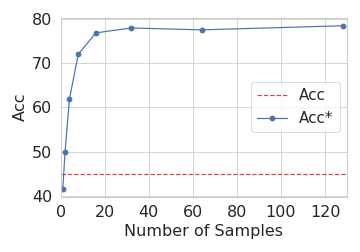}
    }
    \hfill
    \parbox{.48\linewidth}{
    \includegraphics[width=\linewidth]{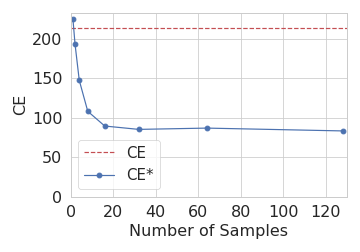}
    }
    \caption{Influence of the number of representation samples used for rectification on Acc (left) and CE (right) for a ResNet-18 trained on CIFAR10. The results are averaged over all severities for Gaussian noise corruption.}
    \label{fig:acc_mse_vs_num_of_samples}
\vspace{-2mm}
\end{figure}

In the preceding experiments, the BN statistics were rectified using only a single batch of $32$ samples. To motivate this hyper-parameter choice, we provide an ablation study analyzing the influence of the number of representation samples on the robustness performance. The results are presented in Figure~\ref{fig:acc_mse_vs_num_of_samples}. A relatively low/high accuracy/mCE is observed using a small number ($1$ to $4$) of representation samples, indicating that the captured statistics are not representative for the overall corruption dataset. A few representation samples as low as $8$ already lead to a sufficiently robust performance above $70\%$. Increasing the number of representation samples lifts the accuracy, with no significant further improvements above $32$ representation samples. Compared with the performance without rectification (dashed line), rectifying with as little as 2 samples already leads to a noticeable performance increase. 

\begin{figure*}[t]
    \centering
    \includegraphics[width=0.9\linewidth]{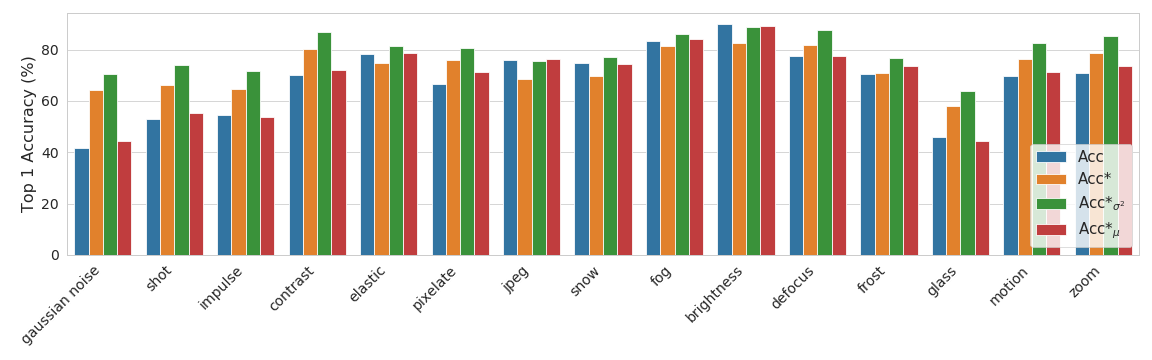}
    \caption{Separate evaluation of corruptions types for a standard ResNet20 trained on Cifar10 (Acc), its accuracy with rectified BN statistics (Acc*), only rectified variance (Acc$_{\sigma^2}$) and only rectified mean (Acc$_{\mu}$). The accuracies are averaged over all 5 severities.}
    \label{fig:comparison}
\end{figure*}

\begin{table}[t]
\centering
\caption{Evaluation of the influence of rectifying the mean $\mu$ or $\sigma^2$ in isolation.  Subscript $\mu$ or $\sigma^2$ indicate that this parameter was rectified.}
\label{tab:mean_var}
\setlength\tabcolsep{2.5pt}
\scalebox{0.9}{
    \begin{tabular}{cc|cc|cc}
    \hline
        Dataset & Model & $\text{Acc*}_{\sigma^2}$ & $\text{Acc*}_{\mu}$ & $\text{mCE*}_{\sigma^2}$ & $\text{mCE*}_{\mu}$\\ \hline
        \multirow{2}{*}{CIFAR10-C}  & ResNet-20 & 79.3 & 69.3 & 69.9 & 102.6 \\ 
                                    & ResNet-56 & 81.6 & 71.7 & 62.7 & 94.7  \\ \hline
        \multirow{2}{*}{CIFAR100-C} & ResNet-20 & 47.3 & 38.9 & 82.9 & 96.2  \\ 
                                    & ResNet-56 & 52.4 & 44.2 & 75.0 & 87.9  \\ \hline
        \multirow{2}{*}{ImageNet-C} & ResNet-18 & 23.4 & 29.0 & 96.6 & 89.2  \\
                                    & ResNet-50 & 34.2 & 37.2 & 83.4 & 78.9  \\
       \hline
    \end{tabular}
    }
\end{table}
\subsection{Impact of mean and variance}
\label{subsec:impact_mean_var}
Rectifying the BN statistics involves the manipulation of two parameters, namely the mean $\mu$ and variance $\sigma^2$.
As an ablation, we study the influence of each parameter in isolation to investigate their contribution to BN rectification. We indicate the rectifiable parameter in the subscript of the metric, \ie $\text{Acc*}_{\sigma^2}$ reports the accuracy for which only the variance ($\sigma^2$) was rectified and the mean ($\mu$) was fixed. The results for the two scenarios are reported in Table~\ref{tab:mean_var}.
For CIFAR rectifying the mean has only a marginal influence. The improvement is never more than $1\%$. Significantly greater improvement is observed when $\mu$ is fixed and only the variance $\sigma^2$ is rectified. For both, ResNet-20 and ResNet-56, with the rectified $\sigma^2$ in most cases a higher accuracy is observed than in the case of rectifying both parameters. For example, under the standard-setting ResNet-20 achieves an Acc* of $73.0\%$, while rectifying only the variance results in an Acc*$_{\sigma^2}$ of $79.3\%$, an additional improvement by $6.3\%$ points.
For ImageNet, however, such a phenomenon can not be observed. Fixing any of the two parameters results in a decrease in accuracy to even lower values than the corruption robustness of the model without rectification.
Overall, we find that for simple datasets like CIFAR10-C and CIFAR100-C, only adapting the variance is sufficient and can even lead to better performances while for a more complex dataset like ImageNet-C, rectifying only one parameter is detrimental and both, $\mu$ and $\sigma^2$ need to be adapted.

Figure~\ref{fig:comparison} breaks down the accuracies before and after rectification as well as with only the rectifiable mean and variance for each corruption type. 
The overall beneficial effect of BN rectification can be observed for most corruption types. Significant improvements can be observed for Gaussian noise, shot, and glass corruption. For the corruptions elastic, jpeg, snow, fog, and brightness the BN rectification slightly decreases the performance. Figure~\ref{fig:comparison} further illustrates the increased performance by only adapting the variance (Acc*$_{\sigma^2}$) and the detrimental effects of rectifying only the mean (Acc*$_{\mu}$). 
However, it is striking that in the cases where BN rectification decreased the performance, rectifying either the mean or the variance parameter results in a better corruption performance than rectifying the parameters in combination, while still achieving only values around the initial corruption robustness.  
Viewing the corruptions individually paints a more nuanced picture and reveals a different interplay between the mean and variance parameters on a case by case basis.

\begin{table}[t]
\centering
\caption{Evaluation of rectifying the BN statistics in different locations of the network. For this purpose, we divide the network into three thirds. All experiments were conducted on CIFAR10-C.}
\setlength\tabcolsep{2.5pt}
    \scalebox{0.9}{
\label{tab:diff_layers_results}
    \begin{tabular}{c|cc|cc|cc}
        \hline
        \multirow{2}{*}{Model} & \multicolumn{2}{c}{Front} & \multicolumn{2}{c}{Middle} & \multicolumn{2}{c}{End}\\
        & Acc* & mCE* & Acc* & mCE* & Acc* & mCE* \\  \hline
        ResNet-20   & 71.7 & 95.7 & 71.5 & 94.7 & 67.2 & 109.3 \\
        ResNet-56   & 80.9 & 64.8 & 73.1 & 89.5 & 70.9 & 97.9 \\
        ResNet-18   & 83.1 & 57.2 & 77.2 & 76.3 & 73.5 & 89.2 \\
        ResNet-50   & 83.7 & 55.5 & 75.6 & 82.0 & 74.5 & 85.4 \\
        VGG-19      & 82.4 & 59.3 & 72.5 & 91.2 & 72.8 & 90.3 \\
        WRN-28-10   & 82.1 & 61.6 & 83.8 & 54.2 & 79.1 & 70.7 \\
        ResNeXt-29  & 83.8 & 56.2 & 78.3 & 74.5 & 76.6 & 79.8 \\
        DenseNet    & 86.0 & 48.2 & 79.2 & 71.5 & 76.9 & 79.9 \\
        \hline
    \end{tabular}
    }
\end{table}

\subsection{Location of rectified parameters}
In the following, we investigate whether the location where the BN statistics will be rectified in the network influences the performance. Therefore, we separate the network into three thirds and only rectify the BN statistics in one of the thirds (front, middle, end). The results are shown in Table~\ref{tab:diff_layers_results}. A trend can be observed that adapting the BN statistics in the first third of the network achieves the highest accuracies.
However, overall most of the performance gains under corruption are worse than the ones where all statistics were adapted. ResNet-50 and VGG-19 are exceptions to this observation, achieving marginal better performance by only rectifying the first third of the BN statistics. 
Rectifying only the BN statistics in the final section leads to accuracies on par with the non-rectified models, indicating only a minor influence on the overall corruption robustness.  

\begin{table*}[t]
\centering
\caption{Comparison of the performance of a ResNet-20 trained on various CIFAF10-C corruptions of severity 3 (Acc$_{\text{UB}}$), the corresponding rectified version of a standard trained model (Acc*) and the accuracy of a standard model on the respective corruption.}
\setlength\tabcolsep{2.5pt}
    \scalebox{0.9}{
    \label{tab:upper_bound}
    \begin{tabular}{ccccccccccccc}
        \hline
        & Gaussian Noise & Shot & Impulse & Defocus & Glass & Motion & Zoom & Snow & Brightness & Contrast & Pixelate & Jpeg \\ \hline
        Acc$_{\text{UB}}$ & 87.55 & 88.42 & 91.46 & 91.70 & 87.55 & 90.38 & 90.90 & 90.04 & 91.59 & 91.57 & 90.70 & 86.37 \\
        Acc* & 62.26 & 63.74 & 66.40 & 83.01 & 62.48 & 75.55 & 78.89 & 70.40 & 82.64 & 80.62 & 77.90 & 68.81 \\
        Acc & 33.41 & 46.31 & 59.17 & 83.13 & 51.22 & 65.85 & 71.75 & 76.68 & 90.44 & 80.56 & 73.06 & 76.14 \\
        \hline
    \end{tabular}
    }
\end{table*}

\subsection{Effectiveness of BN rectification}
By training a model on a training set augmented by a certain corruption, we can achieve a model robust to this particular corruption type. The performance of this model on the respective corruption evaluation dataset can be interpreted as an approximate of the upper bound for this certain corruption type. With such an ``upper bound", indicated by Acc$_{\text{UB}}$ we are able to relate the performance improvement by rectifying the BN statistics. The performances of a ResNet-20 trained on various corruptions for severity 3 is shown in Table~\ref{tab:upper_bound}. As we have already observed in Figure~\ref{fig:comparison}, rectifying the BN statistics does not improve the performance on all corruptions. Consequently, for defocus, snow, brightness and jpeg we also observe a decrease in robustness here. However, it is striking that for these particular corruption types Acc is already relatively close to Acc$_\text{UB}$. For example for brightness, the model without rectified BN statistics exhibits only a gap of around $1\%$ to the performance of a model trained on the corruption. In cases where rectifying the BN statistics leads to an improvement in corruption robustness, a relatively large gap between Acc and Acc$_\text{UB}$ is noticeable. For example for the corruptions of Gaussian noise, shot, impulse glass, and motion the gap between Acc$_\text{UB}$ and Acc* is $25\% \pm 0.5$.

\subsection{Comparison with other normalization techniques}
We evaluate the effect of group normalization (GN) and instance normalization (IN) on the corruption robustness. We train ResNet-18 and ResNet-50 with the respective normalization technique on CIFAR10 and ImageNet and evaluate their corruption robustness. The results for CIFAR10-C and ImageNet-C are shown in Table~\ref{tab:cifar10c_other_normalizations} and Table~\ref{tab:imagenetc_other_normalizations}, respectively. 
For the CIFAR models utilizing GN and IN results in a higher corruption robustness accuracy than for the models with non-rectified BN statistics ($73.9\%$ for ResNet-18 with BN and $74.0\%$ for ResNet-50 with BN). ResNet-50 with IN ($83.2\%$) even outperforms the model with rectified BN statistics ($83.1\%$). 
To evaluate the effect of different normalization on the ImageNet dataset, we train ResNet-18 and ResNet-50 with GN and IN. Similarly to the results in Subsection~\ref{subsec:impact_mean_var}, the results for the ImageNet-C dataset exhibit the opposite trend of the smaller CIFAR10 dataset. With GN the ResNet-18 ($35.1\%$) and ResNet-50 ($43.6\%$) achieve higher corruption robustness compared to the ResNet-18 ($33.7\%$) and ResNet-50 ($40.2\%$) with non-rectified BN statistics. However, the model trained with IN achieves overall the lowest accuracy for ImageNet-C. 

\begin{table}[!htpb]
\centering
\caption{Evaluation of CIFAR10-C on standard models trained on CIFAR-10 with GN and IN.}
\label{tab:cifar10c_other_normalizations}
\scalebox{0.9}{
\begin{tabular}{c|cc|cc}
\hline
\multirow{2}{*}{Model} & \multicolumn{2}{c}{GN} & \multicolumn{2}{c}{IN} \\
        & Acc & mCE & Acc & mCE \\ \hline 
ResNet-18  & 80.5 & 66.3 & 81.2 & 64.2  \\
ResNet-50  & 82.4 & 60.4 & 83.2 & 57.6 \\ \hline
\end{tabular}
}
\end{table}

\begin{table}[!htpb]
\centering
\caption{Evaluation of ImageNet-C on standard models trained on ImageNet with GN and IN.}
\label{tab:imagenetc_other_normalizations}
\scalebox{0.9}{
\begin{tabular}{c|cc|cc}
\hline
\multirow{2}{*}{Model} & \multicolumn{2}{c}{GN} & \multicolumn{2}{c}{IN} \\ 
 & Acc & mCE & Acc & mCE \\ \hline
      
ResNet-18   & 35.1 & 82.2 & 30.0 & 88.8  \\
ResNet-50    & 43.6 & 71.5 & 34.4 & 83.1  \\
\hline
\end{tabular}
}
\end{table}

\subsection{t-SNE analysis}
\begin{figure}[!htbp]
    \centering
    \includegraphics[width=0.95\linewidth]{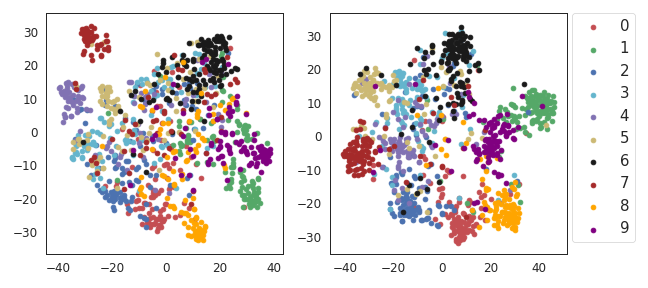}
    \caption{t-SNE for image features of ResNet56 before (left) after (right). $1000$ Gaussian noise corruption images of severity 3 were used.}
    \label{fig:tsne}
\end{figure}

In Figure~\ref{fig:tsne} we visualize the feature vectors of $1000$ randomly chosen images of a model without rectified and rectified BN statistics with t-SNE~\cite{maaten2008visualizing} in 2D. We observe that feature vectors produced by a network with rectified BN statistics are more clustered than the ones resulting from a model without rectified BN statistics.

\subsection{Rectifying BN for adversarial perturbation}
Regarding the model robustness, a parallel line of research analyses adversarial robustness. Small perturbations that are nearly imperceptible to the human eye are able to fool a neural network~\cite{szegedy2013intriguing,goodfellow2014explaining,carlini2017towards,madry2017towards}. Clean samples and adversarial examples have been shown to belong to two different domains~\cite{xie2019intriguing}. Thus, we also experimented with rectifying BN statistics with the representative adversarial examples. However, this technique reduces instead of improving adversarial robustness, which suggests a significant difference between natural corruptions (average-case) and adversarial corruption (worst-case).

\section{Conclusion}
Motivated by the observation that features statistics can be interpreted as the style instead of the content, we proposed to rectify the BN statistics to improve model robustness against common corruptions. Despite simplicity, on several benchmark classification datasets with a wide range of seminal models, we demonstrated that rectifying BN statistics can significantly improve the corruption robustness with an accuracy boost of around $10\%$.We also performed extensive analysis and found that (a) the performance boost increases with the increase of representation samples until it saturates near 32 samples; (b) variance adaptation is sufficient for CIFAR-C, while the more challenging ImageNet-C dataset requires adapting both mean and variance; (c) rectifying the front layers is more crucial than adapting the rear layers; (d) there is a significant imbalance between different corruption regarding the performance boost; (e) other normalization like GN and IN have shown high corruption robustness on CIFAR-C and an opposite trend is observed on ImageNet-C; (f) Rectifying BN statistics also help to make the t-SNE more clustered, which provides some insight on the performance boost; (g) rectifying BN can not help improving adversarial robustness, suggesting a difference between natural corruption and adversarial perturbation.

\section*{Acknowledgment}
This work was supported by the Technology Innovation Program: Development of core technology for advanced locomotion/manipulation based on high-speed/power robot platform and robot intelligence (No.10070171) funded by the Ministry of Trade, Industry \& Energy (MI, Korea)

{\small
\bibliographystyle{ieee_fullname}
\bibliography{bib_mixed}
}

\end{document}